\documentclass[runningheads]{llncs}
\usepackage[T1]{fontenc}
\usepackage{graphicx}
\usepackage{float}
\usepackage{hyperref}
\usepackage{subcaption}
\usepackage{diagbox}
\usepackage{color}

\begin{document}
%

\title{Leveraging Pre-trained Models for Robust Federated Learning for Kidney Stone Type Recognition}
\titlerunning{Federated Learning in Kidney Stone Morpho-constitutional Analysis}
%
\author{Ivan Reyes-Amezcua\inst{1} \and Michael Rojas-Ruiz\inst{2} \and Gilberto Ochoa-Ruiz\inst{3} \and  \\ Andres Mendez-Vazquez\inst{1} \and Christian Daul\inst{4}}

\authorrunning{Reyes et al.}
%
\institute{CINVESTAV, Guadalajara, Mexico \and
Universidad Nacional de Ingenieria, Lima, Peru \and
Tecnologico de Monterrey, School of Engineering and Sciences, Mexico \and CRAN UMR 7039, Université de Lorraine and CNRS, Nancy, France
}
\maketitle              
\begin{abstract}

Deep learning developments have improved medical imaging diagnoses dramatically, increasing accuracy in several domains. Nonetheless, obstacles continue to exist because of the requirement for huge datasets and legal limitations on data exchange. A solution is provided by Federated Learning (FL), which permits decentralized model training while maintaining data privacy. However, FL models are susceptible to data corruption, which may result in performance degradation. Using pre-trained models, this research suggests a strong FL framework to improve kidney stone diagnosis. Two different kidney stone datasets, each with six different categories of images, are used in our experimental setting. Our method involves two stages: Learning Parameter Optimization (LPO) and Federated Robustness Validation (FRV). We achieved a peak accuracy of 84.1\% with seven epochs and 10 rounds during LPO stage, and 77.2\% during FRV stage, showing enhanced diagnostic accuracy and robustness against image corruption. This highlights the potential of merging pre-trained models with FL to address privacy and performance concerns in medical diagnostics, and guarantees improved patient care and enhanced trust in FL-based medical systems.

\keywords{Federated Learning \and Robustness \and Medical Imaging}
\end{abstract}
\section{Introduction}





Recent advancements of artificial intelligence (AI) in medical imaging have transformed healthcare by enhancing diagnostic capabilities and guiding medical procedures across specialized fields including oncology~\cite{bi2019artificial}, neurology~\cite{hamdi2022evaluation}, urology~\cite{Lopez_2021}, and radiology~\cite{ng2021federated}. These innovations have revolutionized the medical practice, facilitating precise diagnosis and customized treatments while also encouraging exploration and research initiatives. However, despite the progress made in establishing large-scale medical datasets, challenges persist, especially in clinical areas such as surgical data science and computer-assisted interventions~\cite{maier2022surgical}. The necessity for extensive datasets for training AI models demands collaborative efforts and data sharing among multiple healthcare institutions~\cite{ali2023multi}. 
Despite these advantages, such collaboration is often constrained by regulatory requirements, including the EU General Data Protection Regulation and other national privacy laws, as well as concerns regarding data privacy and legal implications. To overcome the challenge of privacy concerns, there is a growing research interest in Federated Learning (FL) within the medical field.

FL is a decentralized machine learning (ML) approach where model training occurs locally on data distributed across multiple devices or institutions and model updates occur through central aggregation. This enables collaborative learning while forgoing centralized data sharing, fostering privacy and security regarding the patients' data ~\cite{li2020federated}.
However, FL can encounter unforeseen challenges due to the drifting and heterogeneity in the local datasets, such as image corruptions and perturbations. These perturbations, originating from various sources such as noise or anomalies or even worse, through intentional malicious alterations, can negatively impact the model performance, leading to inaccurate predictions and a loss of trust ~\cite{hendrycks2021natural}. The vulnerability of FL models to this type of corruption underscores the critical importance of developing robust systems, especially in applications such as medical imaging diagnosis. To address the impact of image perturbations on model performance, researchers are actively exploring various strategies, including robust training techniques~\cite{lyu2022privacy}, adversarial training~\cite{hong2022federated}, and the use of pre-trained models~\cite{pmlr-v97-hendrycks19a}. These approaches aim to enhance the resilience of FL models against image corruption by incorporating mechanisms for detecting and mitigating the effects of perturbations during the training process.

To address the challenges posed by corrupted samples from various distributed institutions and the imperative of maintaining privacy in medical imaging, we propose leveraging pre-trained models within a FL framework. 
As highlighted by ~\cite{pmlr-v97-hendrycks19a}, the use of pre-trained models shows promise in enhancing model robustness and uncertainty quantification. Nonetheless, in the context of medical data, safeguarding data privacy remains essential. In this study, we leverage the use of \textit{Flower}, a specialized framework designed for FL~\cite{beutel2020flower}. Flower is an open-source framework that fosters collaboration and innovation in the research community. Its open nature allows for easy sharing, extension, and replication of experiments, promoting transparency and reproducibility in FL research.

Considering these challenges and the potential solutions discussed above, this paper aims to delve deeper into implementing FL with the pre-trained model ResNet18, exploring its efficacy in addressing image corruptions and perturbations to safeguard data privacy in medical imaging within a Flower FL scheme. We are utilizing a medical imaging schema where each federated client is a hospital with different datasets. By examining real applications and experimental results, we aim to shed light on the potential of this approach to revolutionize diagnostic accuracy and patient care in healthcare, particularly focusing on common and critical urological conditions.
To this end, we are using a case study dataset of kidney stone identification across two different data sources. 
This dataset allows us to evaluate the performance and robustness of the FL framework in handling data from diverse origins while ensuring the privacy and security of sensitive medical information. Through this case study, we aim to provide comprehensive insights into the practical applications and benefits of integrating FL with pre-trained models in the medical imaging domain.

The rest of the paper is organized as follows. First, in Section 2 we provide an overview of the state of the art of the FL framework and the challenges related to robustness against image corruption. Section 3 discusses the integration of a pre-trained model, specifically ResNet18, into the FL framework and introduces the datasets used in this research. Additionally, it examines the benefits of this approach for enhancing model robustness through robustness transfer. Section 4 details the experimental setup, including preprocessing methods and the two-stage process of Learning Parameter Optimization (LPO) and Federated Robustness Validation (FRV). In section 5, we analyze the results, focusing on the global model's performance across different scenarios and parameter configurations. Finally, Section 6 concludes the paper by discussing the implications of our findings and suggesting directions for future research.







\section{State of the Art}

\subsection{Federated Learning} FL is an innovative ML technique that enables multiple organizations or institutions to collaboratively solve an ML problem while keeping their data localized \cite{mcmahan2017communication}. Coordinated by a central server, a given model is trained and deployed by distributing it to remote data centers, such as hospitals, allowing each site to maintain its own data privacy. Instead of sharing data, each institution trains the model locally and sends updates back to the central server, which aggregates these updates to improve the global model. This process ensures that data from each contributing center is never exchanged during training, enhancing privacy and security. The iterative process continues until the global model is fully trained.
In FL, an important aspect of the training process is the aggregation of model weights from various institutions. This process can impact the overall model performance, and strategies such as federated averaging (FedAvg), which combines the weights of all models, are commonly used.

A relevant case study can be found in the field of urology, particularly in the analysis and classification of kidney stones using the so-called Morpho-constitutional analysis, particularly using endoscopic images ~\cite{Lopez_2021}. In this context, patient images are confined within individual hospitals, each of which has its restrictions on sharing private data. This scenario presents a significant challenge: developing a robust model despite the presence of corrupted or noisy training images at each hospital. The lack of research on mitigating the impact of noise in kidney stone analysis further underscores the importance of finding effective solutions to enhance model performance in FL frameworks.
To address these challenges, we utilize Flower, a flexible and user-friendly open-source framework designed specifically for FL \cite{beutel2020flower}. 

Flower enables researchers to implement FL systems. Notably, Flower has been identified as a robust framework for FL in medical imaging tasks, as evidenced in the benchmarking study by Fonio \textit{et al.} \cite{fonio2023benchmarking}. This research underscores Flower's scalability and usability, making it a standout choice for healthcare applications.

\subsection{Robustness on Image Corruptions} 
\begin{figure}[t]
    \centering
    \includegraphics[width=1\textwidth]{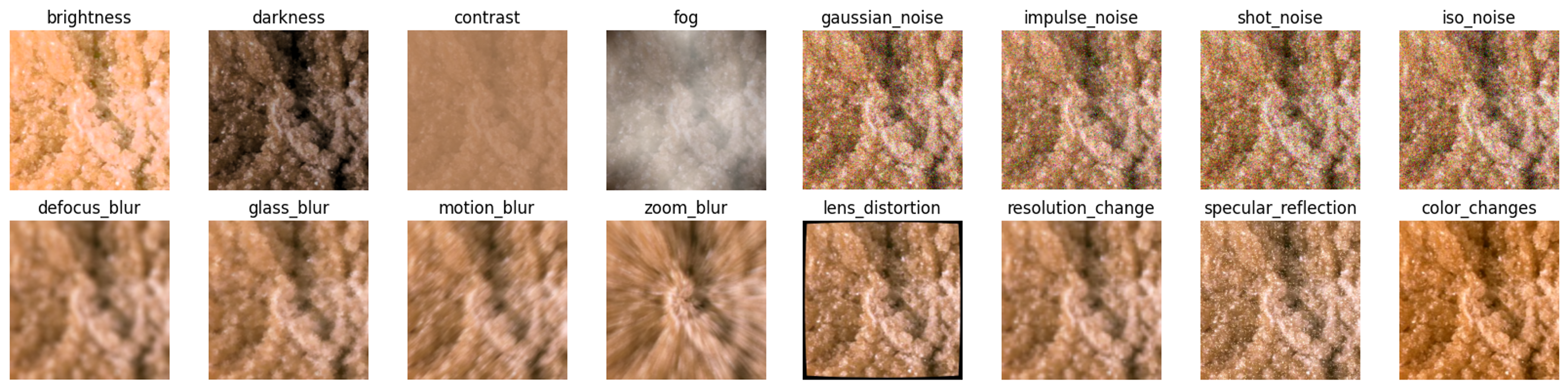}
    \caption{Example of image corruptions on Kidney stone dataset subjected to Section view (SEC) corruption at a third severity level. These image corruption simulates scenarios with structured environmental changes that can affect image quality.}
    \label{fig:sec_corruption}
\end{figure}

Image corruption refers to unintended alterations in image data that can occur due to several factors. Typical examples of corruption in real-world scenarios include Gaussian noise, impulse noise, and defocus blur (Figure \ref{fig:sec_corruption}). Weather conditions also affect image quality, due to the presence of fog, snow, and different levels of brightness can also impact image quality \cite{wang2023survey}. Image corruption is a significant challenge in FL because these perturbations can significantly affect the performance and reliability of the models that are aggregated on the global model. In FL, multiple institutions, such as hospitals, collaborate to train a global model while keeping their data decentralized \cite{sheller2020federated}. However, the diversity in data collection methods, imaging devices, and environmental conditions across these institutions increases the likelihood of encountering corrupted images in FL settings. Recent developments have addressed this issue by employing various strategies to reduce the negative impact of image corruption on performance.

Robustness for computer vision evaluates the susceptibility of deep neural networks (DDNs) to image corruptions. Dan Hendrycks and Thomas Dietterich's \cite{hendrycks2019benchmarking} research introduces ImageNet-C and ImageNet-P benchmarks to assess DNN performance against common real-world corruptions, such as noise and blur. In contrast, Rusak \textit{et al.} \cite{rusak2020simple} study focuses on enhancing model resilience through data augmentation and adversarial training techniques. Additionally, Hendrycks \textit{et al.} \cite{hendrycks2021natural}, introduce the ImageNet-A and ImageNet-O datasets to test model performance against natural adversarial images, further exploring DNN robustness in challenging scenarios. Similarly, in the context of FL, Suyi Li \textit{et al.} \cite{li2020learning} propose a detection framework to identify and exclude malicious clients, defending against adversarial attacks such as Byzantine faults and model poisoning. 

Pre-training models on large, clean datasets and fine-tuning them on smaller, target datasets (under certain conditions) is another effective strategy to improve the model's robustness against image corruption. The effectiveness of this approach has been supported by Kornblith \textit{et al.} \cite{kornblith2019better}, who demonstrated that models with higher performance on ImageNet tend to transfer better to other tasks. Additionally, Hendrycks \textit{et al.} \cite{pmlr-v97-hendrycks19a} found that pre-training not only improves model robustness to adversarial attacks but also enhances uncertainty estimation, making models more reliable in out-of-distribution scenarios. In the context of medical imaging, Raghu \textit{et al.} \cite{raghu2019transfusion} explored the transferability of features from natural images to medical images, providing insights into how pre-training and fine-tuning strategies can be optimized for medical tasks.

\section{Method}

\subsection{Preliminaries}

Kidney stone formation is influenced by factors like diet, low fluid intake, sedentary lifestyle, age, genetics, and chronic diseases. Identifying the types of kidney stones is crucial for appropriate treatment and relapse prevention. The Morpho-Constitutional Analysis (MCA) is the standard method, involving visual inspection and biochemical analysis of kidney stone fragments, but it is time-consuming and requires specialized expertise \cite{daudon2004clinical}. As an alternative, Endoscopic Stone Recognition (ESR) allows for visual identification during ureteroscopy but is subjective and requires significant expertise. To address these limitations, deep-learning methods have been proposed to automate and accelerate kidney stone identification, aiding real-time decision-making during procedures \cite{estrade2017pourquoi}.

Kidney stone morphological and constitutional studies are essential for understanding their biochemical makeup, which is crucial for effective treatment and preventing future episodes \cite{corrales2021classification}. Traditionally, this involves physically extracting stone fragments during ureteroscopy and conducting labor-intensive, time-consuming laboratory analyses. The ability to identify the type of kidney stone directly from in-vivo endoscopic images could transform this process by speeding up diagnosis and potentially eliminating the need for physical extraction, thereby reducing patient exposure to invasive procedures and lowering healthcare costs.

Moreover, the integration of deep learning into kidney stone classification can significantly enhance the accuracy and efficiency of identifying various stone types, which is vital for personalized treatment plans \cite{lopez2023boosting}. Deep learning models, trained on vast datasets of endoscopic images, can recognize subtle patterns and characteristics that may be missed by the human eye, thus providing a more reliable diagnosis.

In the context of medical research and practice, the implementation of federated learning is particularly necessary. Federated learning allows for the development of robust deep learning models across multiple institutions without the need for centralized data sharing, thus addressing privacy concerns associated with patient data \cite{fonio2023benchmarking}. This decentralized approach ensures that sensitive medical information remains local while still contributing to the collective improvement of AI models \cite{chowdhury2023federated}. By harnessing federated learning, medical researchers and practitioners can collaborate effectively, improving diagnostic tools and treatment strategies while maintaining strict patient confidentiality. This is especially critical in healthcare, where data security and privacy are paramount.

\subsection{Federated Learning for Kidney Stone Identification}

\begin{figure}[t]
    \centering
    \includegraphics[width=1\textwidth]{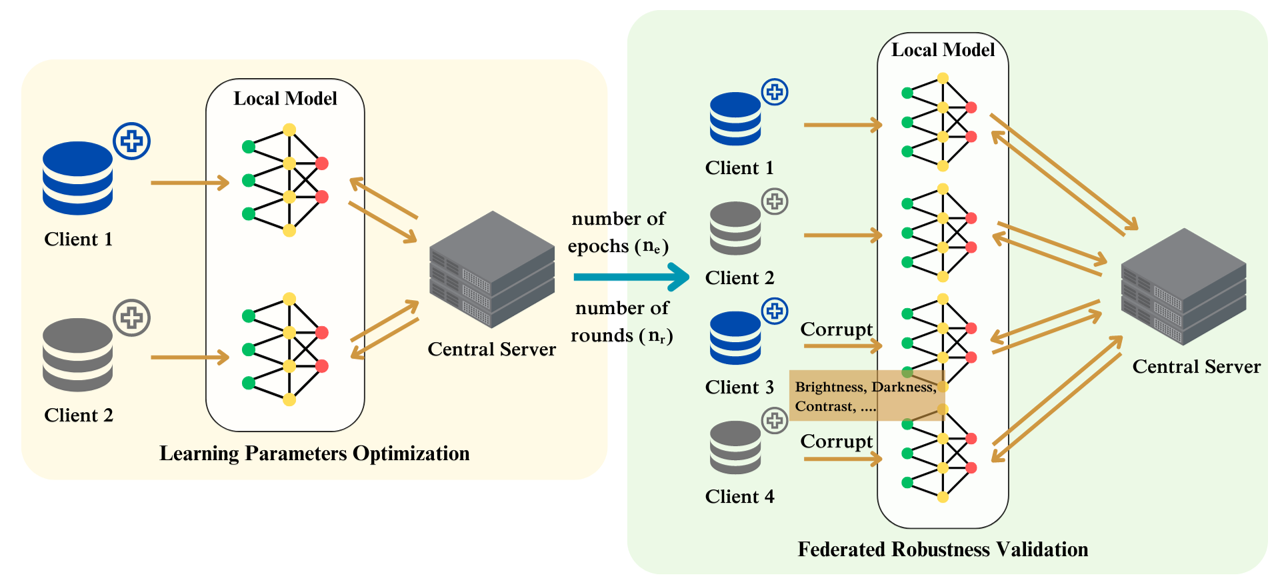}
    \caption{The schematic representation illustrates two stages: Learning Parameters Optimization (LPO), where two clients optimize the $n_e$ and $n_r$ parameters through model training, and Federated Robustness Validation (FRV), which splits datasets into “good” and “corrupted” subsets. The “corrupted” images undergo diverse corruptions to evaluate the model's robustness using the optimized parameters.} 
    \label{fig:illustration}
\end{figure}

In this study, we employ Flower, a Federated Learning framework known for its simplicity and effectiveness in managing complex data distributions and privacy requirements among multiple decentralized nodes (clients). Flower's client-server architecture, illustrated in Figure~\ref{fig:illustration}, allows the server to coordinate and manage the learning process while clients train models locally on their data, preserving privacy. After local training, clients send model updates (weights or gradients) back to the central server, which uses aggregation techniques like FedAvg to refine the global model. This iterative process of training, updating, and aggregating, repeated over several rounds, enhances the model's accuracy, making it a promising approach for real-time kidney stone identification during ureteroscopy.

\textbf{Learning Parameters Optimization (LPO):} The initial stage of this research focuses on determining the optimal number of epochs $(n_e)$ during the local learning phase and the appropriate number of rounds $(n_r)$ for the collaborative learning phase, as depicted in the left part of Figure~\ref{fig:illustration}. These parameters are being identified to establish the conditions under which the FL model reaches its highest accuracy, thereby ensuring optimal performance across distributed settings. To validate our approach, experiments were conducted using two distinct datasets of kidney stones, each representing data from different hospitals. These datasets include six different classes, reflecting the diversity of medical imaging scenarios encountered in each hospital. 

\textbf{Federated Robustness Validation (FRV):} The next stage of the study, shown in the right part of Figure~\ref{fig:illustration}, involves assuming that the parameters $n_e$ and $n_r$, determined in the initial stage, can be applied to the partitions of the datasets used previously. This phase is designed to test the robustness of the FL model in handling noisy clients, such as hospitals, where images may be affected by various types of corruption at different levels of severity.

\subsection{Datasets}

The experiments were conducted using two kidney stone datasets, where the images were captured using either standard CCD cameras or a ureteroscope. Detailed descriptions of these datasets are provided below.

\textbf{Dataset A} \cite{el2022evaluation}: This endoscopic dataset contains 409 images with 246 surface images and 163 section images. It shares the same classes as Dataset A, except Carbapatite fragments are replaced by Weddelite (sub-type IIa) and Uric Acid (IIIa). 

\textbf{Dataset B} \cite{corrales2021classification}: This ex-vivo dataset comprises 366 CCD camera images divided into 209 surface images and 157 section images. It includes six different stone types, categorized by sub-types: WW (Whewellite, subtype Ia), CAR (Carbapatite, IVa), CAR2 (Carbapatite, IVa2), STR (Struvite, IVc), BRU (Brushite, IVd), and CYS (Cystine, Va).

\begin{figure*} [h] 
    \centering
        \subfloat[Dataset A: Jonathan El-Beze]{\label{fig:dataseta}
    \includegraphics[width=0.45\textwidth]{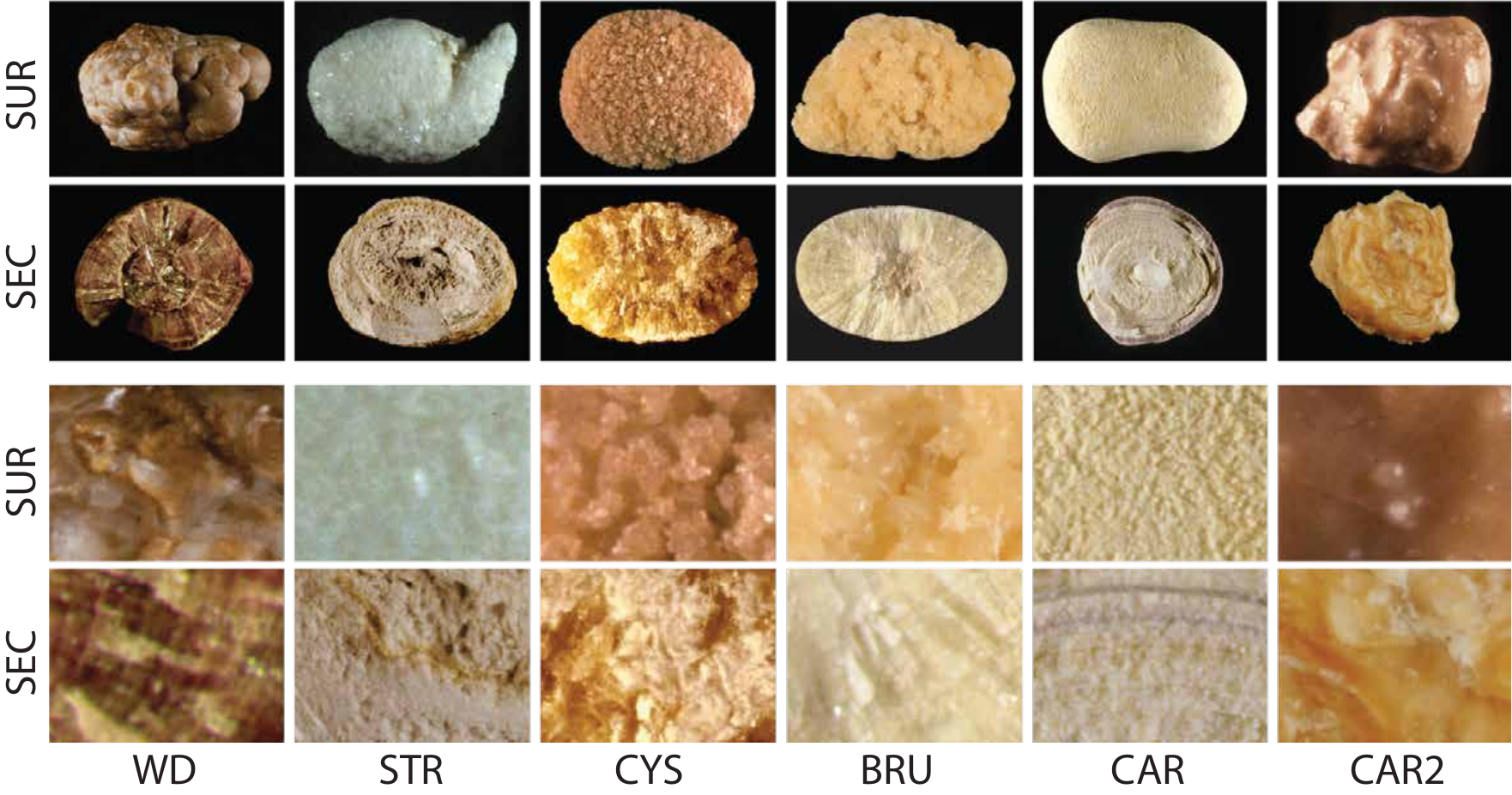}}
    \hspace{5mm}
        \subfloat[Dataset B: Michel Daudon]{
\label{fig:datasetb}\includegraphics[width=0.45\textwidth]{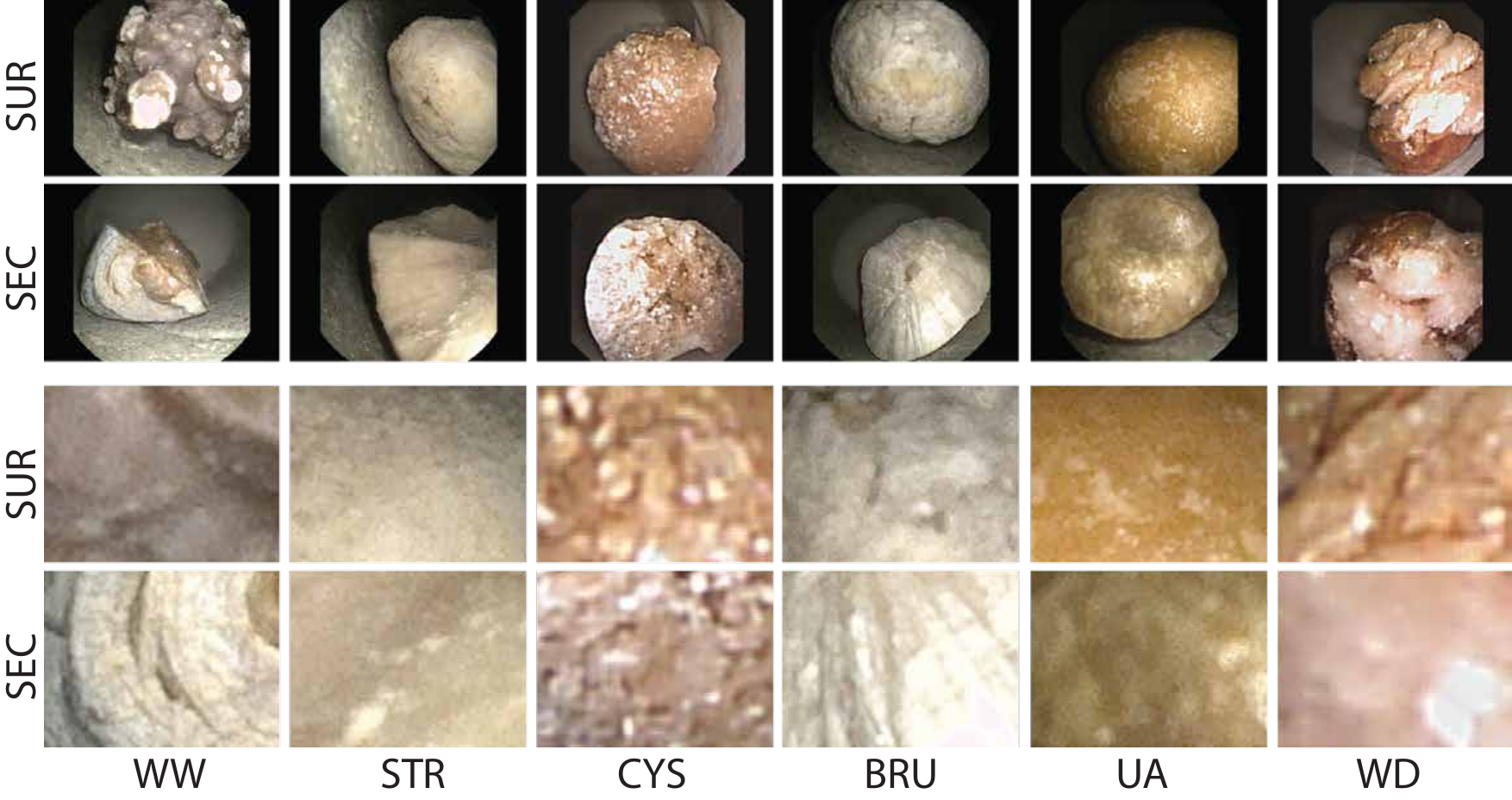}}
    \caption{Samples of ex-vivo kidney stone images captured using (a) a CCD camera and (b) an endoscope. "SEC" refers to section views, and "SUR" refers to surface views.} 
    \label{fig:dataset}
    \end{figure*}

\begin{figure}
    \centering
    \includegraphics[width=0.6\linewidth]{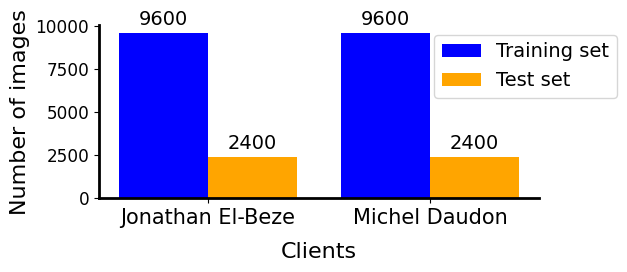}
    \caption{Sizes of the train and test subsets from the Jonathan El-Beze (Dataset A) and Michel Daudon (Dataset B) used in the Learning Parameter Optimization (LPO).}
    \label{fig:two_clients}
\end{figure}

Each dataset generated a total of 12,000 patches, categorized into six classes: Dataset A (WW, STR, CYS, BRU, CAR, CAR2) and Dataset B (WW, WD, UA, STR, BRU, CYS). For both datasets, 80\% of the patches (9,600 patches) are allocated for training and validation, while the remaining 20\% (2,400 patches) are reserved for testing (200 patches per class) (Fig. ~\ref{fig:two_clients}).

\subsection{Robustness Transfer}

In this study, transfer robustness involves using hyperparameters such as the number of epochs and rounds, derived from a model trained under ideal conditions, to guide training in a more challenging environment with corrupted and noisy data \cite{vaishnavi2022transferring,reyes2024enhancing}. This preserves foundational knowledge and enhances performance in adverse conditions.

The process starts with clean training on high-quality images \cite{deng2023hardness}, where the model learns key features and patterns. This phase helps determine optimal hyperparameters for the number of epochs and rounds.

Next, these hyperparameters are applied to training on noisy and corrupted images, crucial for assessing and improving model robustness. This simulates real-world conditions where image quality varies. Using hyperparameters from the clean training phase helps reduce performance loss typically seen with noisy data. This approach is vital in Federated Learning (FL) contexts, where hospitals have varying data-sharing policies and imaging conditions, leading to potential data corruption. Despite these challenges, transfer learning helps maintain model performance and reliability across diverse datasets.

Overall, our FL framework, supported by the Flower platform, combines Federated Robustness Validation (FRV) and Learning Parameters Optimization (LPO) to develop robust models for handling corrupted images, improving diagnostic accuracy and patient care in kidney stone analysis and other medical imaging scenarios.

\section{Experimental Validation}

\subsection{Experimental Setting}


To evaluate the proposed approach, we conducted a series of experiments in two stages. In the first stage, Learning Parameter Optimization (LPO), we utilized different datasets A and B (Fig. \ref{fig:dataset}), each containing data from different hospitals.

All images from each dataset underwent a series of transformations such as cropping, flipping, and normalization to help the model become more adaptable to different image conditions. In contrast, consistent resizing and central cropping during testing ensure fair and reliable performance evaluation. Additionally, before starting the training of the global model, 10\% of the training data was set aside for validation purposes.


The process began with initializing the global model on the server. In the FL framework, the global model parameters were sent to the connected clients (hospitals), ensuring that each participant started their local training with identical model parameters. These parameters were fine-tuned using the weights of ResNet18, pre-trained on ImageNet, and adapted to classify the six different stone types mentioned earlier, enhancing performance accuracy through transfer learning. After local training, each client produced slightly different versions of the model parameters based on their data. These updates were then sent back to the server, where they were combined through a process called aggregation. In this study, Federated Averaging (FedAvg) was employed as the aggregation strategy. FedAvg computes the weighted average of the model updates, with weights based on the number of clients contributing to the training, resulting in a single unified model.

\begin{figure}[h]
    \centering
    \includegraphics[width=1\textwidth]{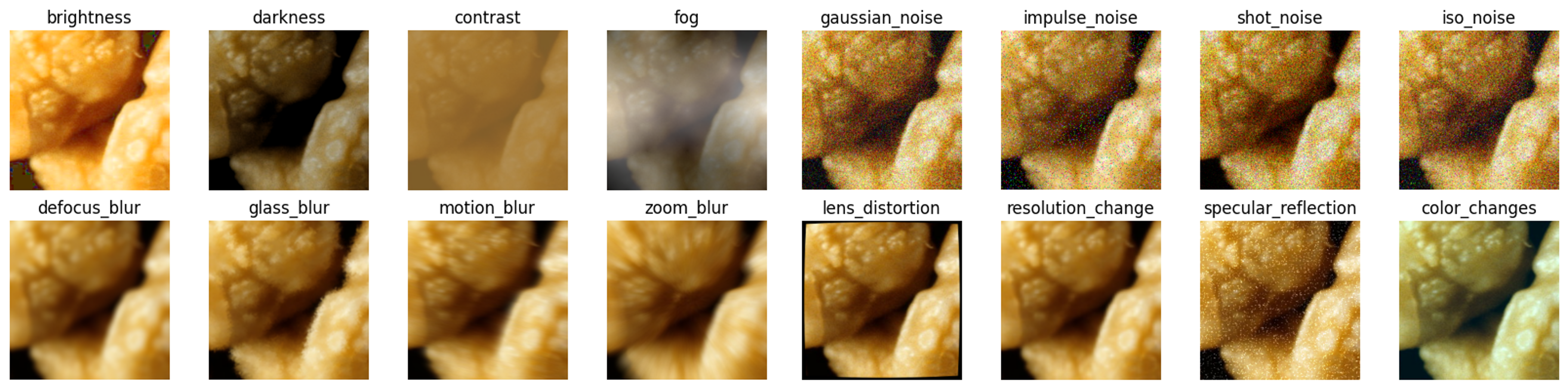}
    \caption{Kidney stone image affected by different types of corruption at the third severity level. This image depicts the MIX subset, showcasing the impact of various image corruptions on kidney stone images from our datasets.}
    \label{fig:mix_corruption}
\end{figure}

The entire process, from the global model parameters being sent to the participating clients to aggregating the updates, constitutes a single round of FL. During each round, the clients only train their local models for a short period. This means that after the aggregation, the model has been trained on the data from all participating clients, but only for a limited time. We then repeat this training process iteratively to eventually arrive at a fully trained model that performs well across the data of all clients.

The LPO stage aims to identify the optimal parameters for the number of epochs ($n_e$) in each local training session and the number of rounds ($n_r$) required for the global model to achieve the highest accuracy. In this stage, we assumed that all clients trained their data for an equal number of epochs. We systematically varied both the number of epochs and rounds, ranging from 1 to 10, to determine the combination that yields the best performance for the global model. Then, these parameters are utilized in the Federated Robustness Validation (FRV) stage. 

\begin{figure}[h]
    \centering
    \includegraphics[width=1\textwidth]{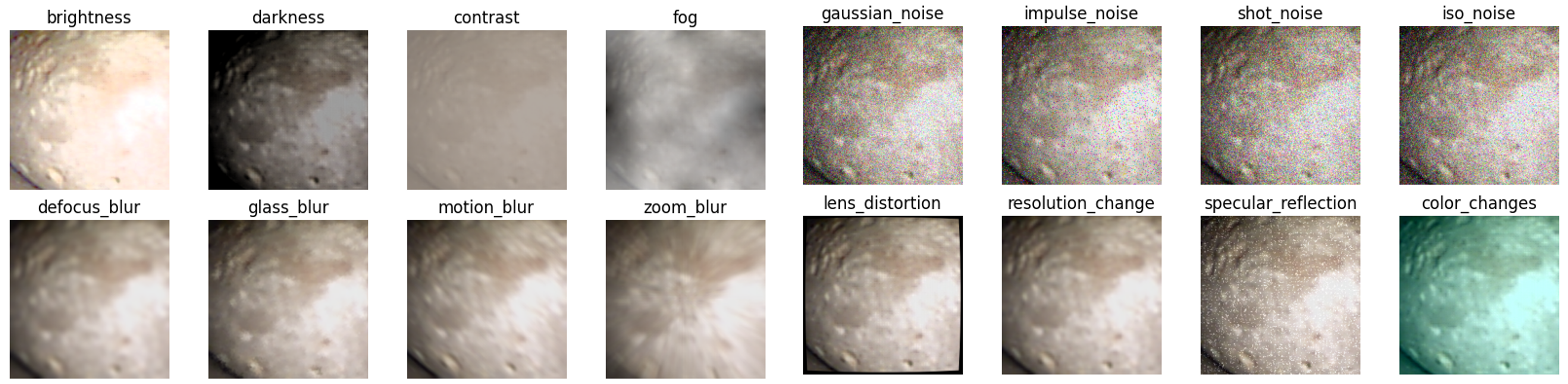}
    \caption{Kidney stone image exposed to various types of corruption at the third severity level. This image represents the SUR subset, demonstrating the effects of different image corruptions on kidney stone visuals from our datasets.}
    \label{fig:sur_corruption}
\end{figure}

\begin{figure}
    \centering
    \includegraphics[width=\linewidth]{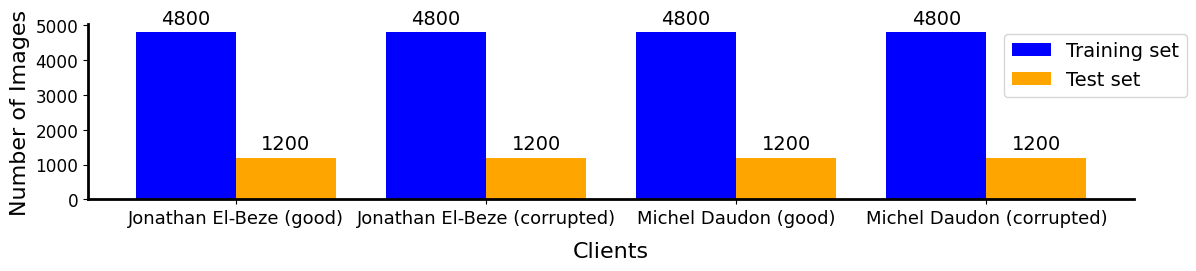}
    \caption{Sizes of the train and test subsets from the four datasets derived by splitting the Jonathan El-Beze and Michel Daudon datasets. Two subsets are labeled as "good" and the other two as "corrupted".}
    \label{fig:four_clients}
\end{figure}

In FRV stage, each dataset is divided equally into two subsets: one half is labeled as “good” and the other half as “corrupted” (Fig.~\ref{fig:four_clients}). This approach is designed to simulate real-world conditions, where data from different hospitals may vary in quality due to diverse factors such as equipment differences, imaging techniques, or environmental conditions. These corruptions are implemented using custom Python scripts that introduce common image corruptions such as changes in brightness, darkness, contrast variations, motion blur, etc. into clean images. Therefore, the “corrupted” subsets include images intentionally subjected to various types of corruption at different severity levels (Fig. \ref{fig:sec_corruption}, \ref{fig:mix_corruption}, \ref{fig:sur_corruption}). Corrupted images are generated by randomly sampling the corruption types and severity levels from a uniform distribution.

Before initializing the global model on the server, the images from the four clients underwent the same series of transformations as in the first stage. Assuming the optimal parameters identified above are applicable in a four-client setting, they are applied in this FL framework for both the local training of each client and the training of the global model. Similar to the LPO stage, the global model parameters were fine-tuned using ResNet18 weights, pre-trained on ImageNet. FedAvg was also employed as the aggregation strategy.

The objective of this stage is to evaluate the robustness of the global model by applying the optimal parameters for both local and global training in a real-world setting, which includes handling image corruption across different clients while preserving data privacy.

Both the LPO and FRV stages operate ensuring that no data is exchanged between clients, which is a paramount feature of an FL framework. Additionally, the Adam optimizer was used in this study with a learning rate of $\alpha = 0.0001$ and a weight decay of $1 \times 10^{-5}$. The training was conducted with a batch size of 4, and the Cross-Entropy Loss function was employed during both the training and testing phases. All experiments were implemented in Python 3.10 using PyTorch 2.3.0+cu121 and the Flower 1.9.0 framework. The model was trained on a Tesla T4 GPU (15 GB) provided by Google Colaboratory.

\section{Results}

Table~\ref{tab:LPO_results} illustrates the performance of the FL model across 10 epochs and 10 rounds during the \textbf{LPO} stage. The heatmap reveals that the highest accuracy, \textbf{84.1\%}, is achieved using 7 rounds and 10 epochs.



\begin{table}[h]
    \centering
    \begin{tabular}{|c||*{10}{c|}}
    \hline
    \backslashbox{\scriptsize Number\\of \\epochs ($n_e$)}{\scriptsize Number\\of \\rounds ($n_r$)} & \scriptsize \makebox[2em]{1} & \scriptsize \makebox[2em]{2} & \scriptsize \makebox[2em]{3} & \scriptsize \makebox[2em]{4} & \scriptsize \makebox[2em]{5} & \scriptsize \makebox[2em]{6} & \scriptsize \makebox[2em]{7} & \scriptsize \makebox[2em]{8} & \scriptsize \makebox[2em]{9} & \scriptsize \makebox[2em]{10} \\ \hline\hline
     \scriptsize 1 & \scriptsize 0.671 & \scriptsize 0.679 & \scriptsize 0.681 & \scriptsize 0.704 & \scriptsize 0.770 & \scriptsize 0.680 & \scriptsize 0.710 & \scriptsize 0.737 & \scriptsize 0.785 & \scriptsize 0.777 \\ \hline
     \scriptsize 2 & \scriptsize 0.695 & \scriptsize 0.713 & \scriptsize 0.754 & \scriptsize 0.743 & \scriptsize 0.794 & \scriptsize 0.770 & \scriptsize 0.774 & \scriptsize 0.734 & \scriptsize 0.774 & \scriptsize 0.745 \\ \hline
     \scriptsize 3 & \scriptsize 0.661 & \scriptsize 0.703 & \scriptsize 0.727 & \scriptsize 0.760 & \scriptsize 0.782 & \scriptsize 0.780 & \scriptsize 0.769 & \scriptsize 0.771 & \scriptsize 0.800 & \scriptsize 0.815 \\ \hline
     \scriptsize 4 & \scriptsize 0.683 & \scriptsize 0.705 & \scriptsize 0.739 & \scriptsize 0.768 & \scriptsize 0.738 & \scriptsize 0.735 & \scriptsize 0.772 & \scriptsize 0.797 & \scriptsize 0.782 & \scriptsize 0.807 \\ \hline
     \scriptsize 5 & \scriptsize 0.634 & \scriptsize 0.683 & \scriptsize 0.701 & \scriptsize 0.702 & \scriptsize 0.739 & \scriptsize 0.783 & \scriptsize 0.818 & \scriptsize 0.791 & \scriptsize 0.810 & \scriptsize 0.816 \\ \hline
     \scriptsize 6 & \scriptsize 0.618 & \scriptsize 0.618 & \scriptsize 0.670 & \scriptsize 0.655 & \scriptsize 0.702 & \scriptsize 0.725 & \scriptsize 0.748 & \scriptsize 0.771 & \scriptsize 0.794 & \scriptsize 0.818 \\ \hline
     \scriptsize 7 & \scriptsize 0.656 & \scriptsize 0.683 & \scriptsize 0.709 & \scriptsize 0.698 & \scriptsize 0.734 & \scriptsize 0.770 & \scriptsize 0.737 & \scriptsize 0.773 & \scriptsize 0.809 & \scriptsize \textbf{0.841} \\ \hline
     \scriptsize 8 & \scriptsize 0.540 & \scriptsize 0.681 & \scriptsize 0.763 & \scriptsize 0.772 & \scriptsize 0.781 & \scriptsize 0.790 & \scriptsize 0.800 & \scriptsize 0.809 & \scriptsize 0.818 & \scriptsize 0.828 \\ \hline
     \scriptsize 9 & \scriptsize 0.602 & \scriptsize 0.702 & \scriptsize 0.736 & \scriptsize 0.748 & \scriptsize 0.759 & \scriptsize 0.770 & \scriptsize 0.781 & \scriptsize 0.793 & \scriptsize 0.804 & \scriptsize 0.815 \\ \hline
     \scriptsize 10 & \scriptsize 0.655 & \scriptsize 0.614 & \scriptsize 0.618 & \scriptsize 0.623 & \scriptsize 0.627 & \scriptsize 0.632 & \scriptsize 0.636 & \scriptsize 0.641 & \scriptsize 0.646 & \scriptsize 0.649 \\ \hline
    \end{tabular}
    \caption{Accuracy achieved by the global model over 10 epochs and 10 rounds during the Learning Parameter Optimization (LPO) stage.}
    \label{tab:LPO_results}
\end{table}


This configuration is used in the next stage, \textbf{FRV}, where corrupted clients are included to assess the model’s resilience under more challenging conditions. Despite the introduction of data corruptions that simulate real-world disruptions encountered in clinical environments, the model maintains a commendable accuracy of \textbf{77.2\%}.

\section{Conclusions}

In this work, we applied FL with the Flower framework to address the problems caused by noisy and distorted images in medical imaging, particularly for urology's kidney stone analysis. We used two stages in our methodology: FRV using noisy and corrupted images to imitate real-world settings, and LPO utilizing clean datasets to build a solid model foundation. Through the application of transfer learning, we improved the robustness and resilience of the model by using pre-trained weights from the clean model to reduce performance deterioration when training with noisy data.

The findings show that when clean initial training is combined with later robust validation, FL models' accuracy and dependability when handling a variety of decentralized datasets are much increased. In the field of medical imaging, where data privacy and quality vary throughout institutions, this technique is very beneficial. The Flower framework highlighted the potential of FL to improve patient care and diagnostic accuracy in healthcare by enabling collaborative learning across several hospitals while maintaining data privacy. Subsequent investigations will concentrate on enhancing the robustness of the model and investigating supplementary medical imaging uses to promote decentralized and privacy-preserving machine learning in healthcare.

\section*{Acknowledgments}
The authors wish to acknowledge the Mexican Council for Science and Technology (CONAHCYT) for their support in terms of postgraduate scholarships in this project, and the Data Science Hub at Tecnologico de Monterrey for their support on this project. 
This work has been supported by Azure Sponsorship credits granted by Microsoft's AI for Good Research Lab through the AI for Health program. The project was also supported by the French-Mexican ANUIES CONAHCYT Ecos Nord grant 322537. Finally, Michael Rojas work was supported by the Google ExploreCSR project ``LATAM Undergraduate Research Program (2024)." 

\bibliographystyle{splncs04}
\bibliography{sn-bibliography}

\begin{thebibliography}{10}
\providecommand{\url}[1]{\texttt{#1}}
\providecommand{\urlprefix}{URL }
\providecommand{\doi}[1]{https://doi.org/#1}

\bibitem{ali2023multi}
Ali, S., Jha, D., Ghatwary, N., Realdon, S., Cannizzaro, R., Salem, O.E., Lamarque, D., Daul, C., Riegler, M.A., Anonsen, K.V., et~al.: A multi-centre polyp detection and segmentation dataset for generalisability assessment. Scientific Data  \textbf{10}, ~75 (2023). \doi{10.1038/s41597-023-01981-y}

\bibitem{beutel2020flower}
Beutel, D.J., Topal, T., Mathur, A., Qiu, X., Fernandez-Marques, J., Gao, Y., Sani, L., Li, K.H., Parcollet, T., de~Gusm{\~a}o, P.P.B., et~al.: Flower: A friendly federated learning research framework (2020), preprint at \url{https://arxiv.org/abs/2007.14390}

\bibitem{bi2019artificial}
Bi, W.L., Hosny, A., Schabath, M.B., Giger, M.L., Birkbak, N.J., Mehrtash, A., Allison, T., Arnaout, O., Abbosh, C., Dunn, I.F., et~al.: Artificial intelligence in cancer imaging: clinical challenges and applications. CA: a cancer journal for clinicians  \textbf{69},  127--157 (2019). \doi{10.3322/caac.21552}

\bibitem{chowdhury2023federated}
Chowdhury, D., Banerjee, S., Sannigrahi, M., Chakraborty, A., Das, A., Dey, A., Dwivedi, A.D.: Federated learning based covid-19 detection. Expert Systems  \textbf{40},  e13173 (2023). \doi{10.1111/exsy.13173}

\bibitem{corrales2021classification}
Corrales, M., Doizi, S., Barghouthy, Y., Traxer, O., Daudon, M.: Classification of stones according to michel daudon: a narrative review. European Urology Focus  \textbf{7}(1),  13--21 (2021)

\bibitem{daudon2004clinical}
Daudon, M., Jungers, P.: Clinical value of crystalluria and quantitative morphoconstitutional analysis of urinary calculi. Nephron Physiology  \textbf{98}(2),  p31--p36 (2004)

\bibitem{deng2023hardness}
Deng, Y., Gazagnadou, N., Hong, J., Mahdavi, M., Lyu, L.: On the hardness of robustness transfer: A perspective from rademacher complexity over symmetric difference hypothesis space (2023), preprint at \url{https://arxiv.org/abs/2302.12351}

\bibitem{el2022evaluation}
El~Beze, J., Mazeaud, C., Daul, C., Ochoa-Ruiz, G., Daudon, M., Eschw{\`e}ge, P., Hubert, J.: Evaluation and understanding of automated urinary stone recognition methods. BJU international  (2022)

\bibitem{estrade2017pourquoi}
Estrade, V., Daudon, M., Traxer, O., M{\'e}ria, P., et~al.: Pourquoi l’urologue doit savoir reconna{\^\i}tre un calcul et comment faire? les bases de la reconnaissance endoscopique. Progr{\`e}s en Urologie-FMC  \textbf{27}(2),  F26--F35 (2017)

\bibitem{fonio2023benchmarking}
Fonio, S.: Benchmarking federated learning frameworks for medical imaging tasks (2024), paper presented at Image Analysis and Processing - ICIAP 2023 Workshops, 21 January 2024

\bibitem{hamdi2022evaluation}
Hamdi, M., Bourouis, S., Rastislav, K., Mohmed, F.: Evaluation of neuro images for the diagnosis of alzheimer's disease using deep learning neural network. Frontiers in Public Health  \textbf{10},  834032 (2022). \doi{10.3389/fpubh.2022.834032}

\bibitem{hendrycks2019benchmarking}
Hendrycks, D., Dietterich, T.: Benchmarking neural network robustness to common corruptions and perturbations (2019), preprint at \url{https://arxiv.org/abs/1903.12261}

\bibitem{pmlr-v97-hendrycks19a}
Hendrycks, D., Lee, K., Mazeika, M.: Using pre-training can improve model robustness and uncertainty (2019), preprint at \url{https://arxiv.org/abs/1901.09960}

\bibitem{hendrycks2021natural}
Hendrycks, D., Zhao, K., Basart, S., Steinhardt, J., Song, D.: Natural adversarial examples (2021), preprint at \url{https://arxiv.org/abs/1907.07174}

\bibitem{hong2022federated}
Hong, J., Wang, H., Wang, Z., Zhou, J.: Federated robustness propagation: Sharing adversarial robustness in federated learning (2022), preprint at \url{https://arxiv.org/abs/2106.10196}

\bibitem{kornblith2019better}
Kornblith, S., Shlens, J., Le, Q.V.: Do better imagenet models transfer better? (2019), preprint at \url{https://arxiv.org/abs/1805.08974}

\bibitem{li2020learning}
Li, S., Cheng, Y., Wang, W., Liu, Y., Chen, T.: Learning to detect malicious clients for robust federated learning  (2020), preprint at \url{https://arxiv.org/abs/2002.00211}

\bibitem{li2020federated}
Li, T., Sahu, A.K., Talwalkar, A., Smith, V.: Federated learning: Challenges, methods, and future directions. IEEE signal processing magazine  \textbf{37},  50--60 (2020). \doi{10.1109/MSP.2020.2975749}

\bibitem{Lopez_2021}
Lopez, F., Varelo, A., Hinojosa, O., Mendez, M., Trinh, D.H., ElBeze, Y., Hubert, J., Estrade, V., Gonzalez, M., Ochoa, G., Daul, C.: Assessing deep learning methods for the identification of kidney stones in endoscopic images (2021), paper presented at the 43rd Annual International Conference of the IEEE Engineering in Medicine and Biology Conference (EMBC), November 2021

\bibitem{lopez2023boosting}
Lopez-Tiro, F., Flores-Araiza, D., Betancur-Rengifo, J.P., Reyes-Amezcua, I., Hubert, J., Ochoa-Ruiz, G., Daul, C.: Boosting kidney stone identification in endoscopic images using two-step transfer learning. In: Mexican International Conference on Artificial Intelligence. pp. 131--141. Springer (2023)

\bibitem{lyu2022privacy}
Lyu, L., Yu, H., Ma, X., Chen, C., Sun, L., Zhao, J., Yang, Q., Philip, S.Y.: Privacy and robustness in federated learning: Attacks and defenses. IEEE transactions on neural networks and learning systems pp. 1--21 (2022). \doi{10.1109/TNNLS.2022.3216981}

\bibitem{maier2022surgical}
Maier-Hein, L., Eisenmann, M., Sarikaya, D., M{\"a}rz, K., Collins, T., Malpani, A., Fallert, J., Feussner, H., Giannarou, S., Mascagni, P., et~al.: Surgical data science--from concepts toward clinical translation. Medical image analysis  \textbf{76},  102306 (2022). \doi{10.1016/j.media.2021.102306}

\bibitem{mcmahan2017communication}
McMahan, B., Moore, E., Ramage, D., Hampson, S., y~Arcas, B.A.: Communication-efficient learning of deep networks from decentralized data (2017), preprint at \url{https://arxiv.org/abs/1602.05629}

\bibitem{ng2021federated}
Ng, D., Lan, X., Yao, M.M.S., Chan, W.P., Feng, M.: Federated learning: a collaborative effort to achieve better medical imaging models for individual sites that have small labelled datasets. Quantitative Imaging in Medicine and Surgery  \textbf{11}, ~852 (2021). \doi{10.21037/qims-20-595}

\bibitem{raghu2019transfusion}
Raghu, M., Zhang, C., Kleinberg, J., Bengio, S.: Transfusion: Understanding transfer learning for medical imaging (2019), preprint at \url{https://arxiv.org/abs/1902.07208}

\bibitem{reyes2024enhancing}
Reyes-Amezcua, I., Ochoa-Ruiz, G., Mendez-Vazquez, A.: Enhancing image classification robustness through adversarial sampling with delta data augmentation (dda) (2024), paper presented at the IEEE/CVF Conference on Computer Vision and Pattern Recognition

\bibitem{rusak2020simple}
Rusak, E., Schott, L., Zimmermann, R.S., Bitterwolf, J., Bringmann, O., Bethge, M., Brendel, W.: A simple way to make neural networks robust against diverse image corruptions. In: Proceedings of the Computer Vision--ECCV 2020: 16th European Conference, Glasgow, UK, August 23--28, 2020, Part III 16. pp. 53--69 (2020)

\bibitem{sheller2020federated}
Sheller, M.J., Edwards, B., Reina, G.A., Martin, J., Pati, S., Kotrotsou, A., Milchenko, M., Xu, W., Marcus, D., Colen, R.R., et~al.: Federated learning in medicine: facilitating multi-institutional collaborations without sharing patient data. Scientific reports  \textbf{10},  12598 (2020). \doi{10.1038/s41598-020-69250-1}

\bibitem{vaishnavi2022transferring}
Vaishnavi, P., Eykholt, K., Rahmati, A.: Transferring adversarial robustness through robust representation matching (2022), paper presented at the 31st USENIX Security Symposium (USENIX Security 22)

\bibitem{wang2023survey}
Wang, S., Veldhuis, R., Brune, C., Strisciuglio, N.: A survey on the robustness of computer vision models against common corruptions (2023), preprint at \url{https://arxiv.org/abs/2305.06024}

\end{thebibliography}

\end{document}